\documentclass{article}

\usepackage[square, numbers]{natbib}

\usepackage[preprint]{neurips_fitml_2024}

\usepackage[utf8]{inputenc} 
\usepackage[T1]{fontenc}    
\usepackage{hyperref}       
\usepackage{url}            
\usepackage{booktabs}       
\usepackage{multirow}
\usepackage{amsfonts}       
\usepackage{amssymb}        
\usepackage{amsmath}        
\usepackage{nicefrac}       
\usepackage{microtype}      
\usepackage{xcolor}         
\usepackage{graphicx}       
\usepackage{todonotes}      
\usepackage{subcaption}     
\usepackage[capitalize]{cleveref}
\usepackage{wrapfig}        
\usepackage[title]{appendix}

\newcommand{{\ourmethod}}{GTX}

\title{Fine-tuning Vision Classifiers On A Budget}

\author{%
  Sunil Kumar\\    
  Groundlight.ai\\
  Seattle, WA 98122 \\
  \texttt{sunil@groundlight.ai} \\
  \And
  Ted Sandler\\
  Groundlight.ai \\
  Seattle, WA 98122 \\
  \texttt{ted@groundlight.ai} \\
  \And 
  Paulina Varshavskaya \\
  Groundlight.ai \\
  Seattle, WA 98122 \\
  \texttt{paulina@groundlight.ai} \\
}

\begin{document}

\maketitle

\begin{abstract}
  Fine-tuning modern computer vision models requires accurately labeled data for which the ground truth may not exist, but a set of multiple labels can be obtained from labelers of variable accuracy. We tie the notion of label quality to confidence in labeler accuracy and show that, when prior estimates of labeler accuracy are available, using a simple naive-Bayes model to estimate the true labels allows us to label more data on a fixed budget without compromising label or fine-tuning quality. We present experiments on a dataset of industrial images  that demonstrates that our method, called Ground Truth Extension (\ourmethod), enables fine-tuning ML models using fewer human labels.
\end{abstract}

\begin{wrapfigure}{r}{0.5\textwidth}
    \centering
    \includegraphics[width=0.48\textwidth,height=6cm]{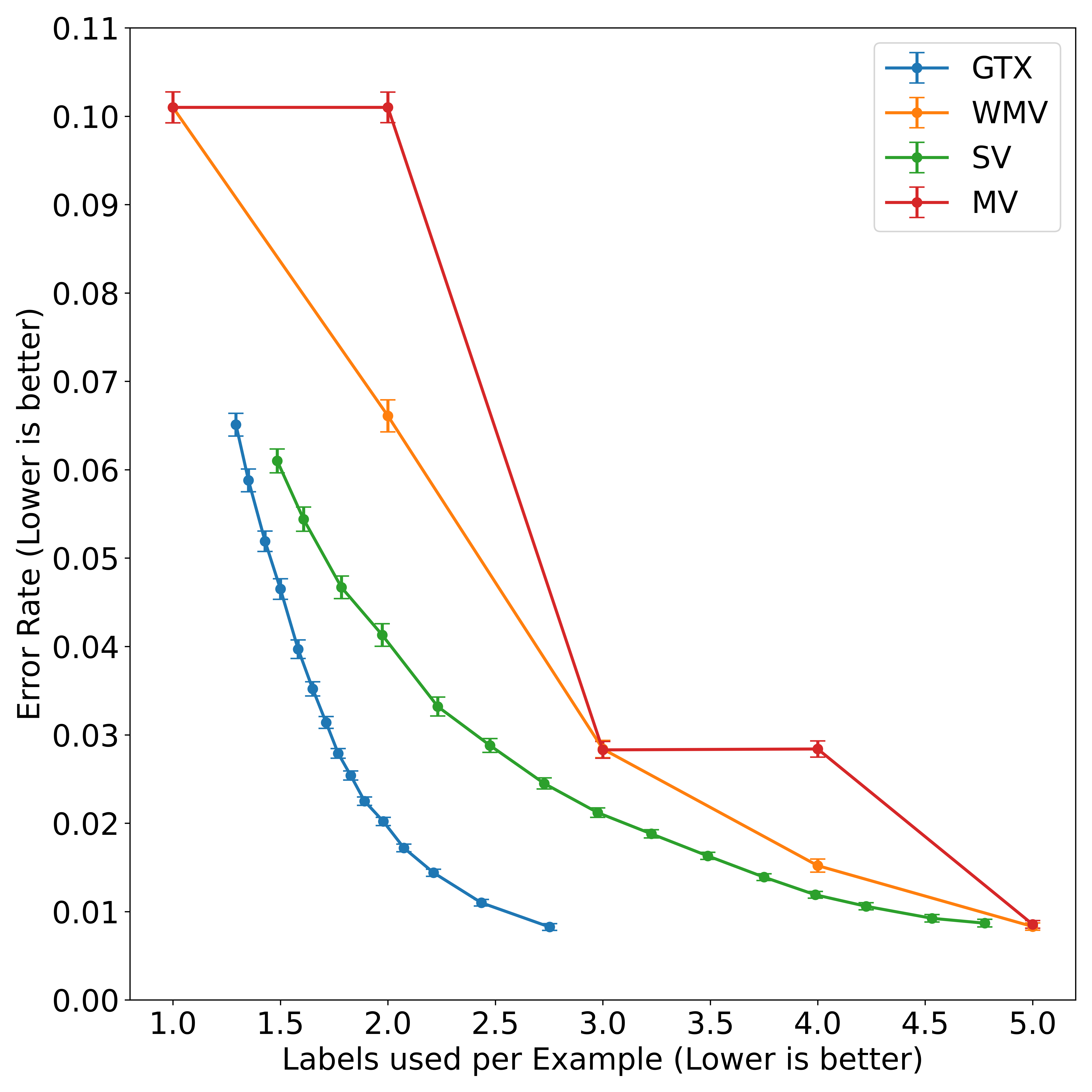}
    \caption{\textbf{Efficiency of Label Aggregation Methods.} The figure shows the error rate of various label aggregation methods as a function of the number of labels per example. GTX demonstrates superior efficiency by using fewer labels to achieve lower error rates. See Section 3.4 for more details.}
    \label{fig:c_t_experiment_pareto_line_graph}
\end{wrapfigure}

\section{Introduction}

Fine-tuning modern ML models requires large quantities of high-quality, accurately labeled and relevant data. What counts as high quality? The two standard approaches for any specific dataset labeling task are either to engage trustworthy experts and treat their labels as ground truth, or crowd-source multiple labels of variable quality for each data point and aggregate them to impute trustworthy labels. 

The quality of the aggregation result matters. A dataset with a significant presence of label noise from variable-quality human labels is problematic, especially for large models that can easily overfit to labeling errors. This is specifically true in computer vision where deep learning models have been shown to fit even random labels \citep{zhang2021}. Learning in the presence of noisy labels has been addressed in the literature in a number of ways from direct dataset cleaning to regularization and ensemble methods \citep{li2024}. The cost of label collection also matters. For a fixed labeling budget, minimizing the number of potentially noisy labels required to estimate the true label of a data point makes the labeling budget go further in terms of the total number of examples whose true labels we can infer.

So what is (ground) truth anyway? What constitutes a trustworthy or accurate label in the absence of expert opinion? In this paper, we tie label quality to a confidence estimated from historical labeler accuracy and show that using the one-coin naive-Bayes response model described in \cite{dawidskene} to estimate this confidence allows us to label more data on a fixed budget without compromising label quality or the quality of the fine-tuned model. Our focus is on optimizing a finite labeling budget to get the most truth and best quality data for a more label-efficient fine-tuning. This focus is complementary to the active learning question of which examples to select for labeling \citep{settles2009}.

The authors of \cite{dawidskene} focus on the scenario in which no true labels are available, and therefore they utilize the EM algorithm to estimate both the true labels and the labeler accuracies. However, in many settings, one has access to some labeler accuracy information. For instance, it is often possible to collect at least a small number of expertly labeled examples which we call ground truth labels, and to evaluate the non-expert labelers on these. Or alternatively, we may have observed labelers' performances on previous labeling tasks for which ground truth labels were available. In these cases, we can run inference with the naive-Bayes model to infer the true labels (the E-step) and forego the M-step to re-estimate labeler accuracies. This is our method which we call Ground Truth Extension (\ourmethod). 

Therefore, in this paper, we consider a scenario where labelers come from a known and limited workforce cohort, such that we can estimate their individual labeling accuracies, but do not have the ability to route specific data points to particular labelers, as is common when using online data labeling services such as LabelBox\footnote{https://docs.labelbox.com/docs/quality-analysis\#view-and-assess-quality-analysis-performance} or in vertically integrated human-in-the-loop ML services such as Groundlight.ai. We also do not consider large-pool crowdsourcing marketplace-based services such as Amazon Mechanical Turk, where we may not have access to labeler accuracy statistics on prior labeling tasks. 

We demonstrate across a variety of circumstances that the \ourmethod\ method of imputing ground truth from multiple labels of variable quality, which we describe in section \ref{sec: method}, outperforms standard and weighted majority voting in accuracy of imputed labels. We show this in a series of controlled synthetic data experiments (section \ref{sec: synthetic_experiments}), as well as in an industrial visual classification task where \ourmethod-inferred labels are used to fine-tune a CNN vision model (section \ref{sec: real_data_experiments}).

\section{Ground Truth Extension} \label{sec: method}

The \ourmethod\ method is a framework for decision-making about label trustworthiness. It can facilitate more data to be labeled accurately on a constrained budget and consists of three parts: 
\begin{enumerate}
\item A probabilistic model for estimating the true label and its confidence given noisy labels from non-expert labelers and estimates of their accuracies.
\item A method to estimate labeler accuracies from a comparatively small set of labels provided by experts -- \emph{ground truth labels}.
\item A strategy for determining which examples to (noisily) label in order to maximize the number of examples whose true labels can be inferred at a given level of confidence.
\end{enumerate}

\subsection{Naive-Bayes model of true labels}
\label{subsec: imputing_gt}

We use the one-coin naive-Bayes model from \cite{dawidskene, ok2016, Khetan2017} to infer a true label estimate and our confidence in its correctness from a collection of noisy labels. This model assumes that each labeler acts independently and has a fixed, but unknown, level of accuracy that is independent of the true label. The label confidence is taken to be the probability of that label under the one-coin probabilistic model.

Let $y_i \in \{0,1\}$ denote the true label of example $i$ and let $y_i^j$ be the label that labeler $j \in \mathcal{J}$ assigns to example $i$. Since not all labelers label each instance, we use $\mathcal{J}(i)$ to denote the set of labelers who labeled example $i$ and $\mathcal{L}(i) = \{ y_i^j \mid j \in \mathcal{J}(i) \}$ to be the set of noisy labels they provided. 

In the model, the probability that a randomly selected example $i$ has true label $y$ is $\pi_y = P(Y_i = y)$, and the probability that labeler $j$ assigns the correct label to example $i$ is $\alpha_j = P(Y_i^j = y_i)$ for all $i$. 
The labelers' responses are assumed independent and therefore
\begin{eqnarray}\label{eq:indep}
P( \mathcal{L}(i) \mid Y_i = y )
= \prod_{j \in \mathcal{J}(i)} \alpha_j^{\mathbf{I}[y_i^j = y]} \, (1 - \alpha_j)^{\mathbf{I}[y_i^j \not= y]},
\end{eqnarray}
where $\mathbf{I}[\cdot]$ is the indicator function that equals 1 if its argument is true and 0 otherwise.

By Bayes rule, the probability that example $i$ has true label $y$ is
\begin{eqnarray}
P( Y_i = y \mid \mathcal{L}(i) )
&\propto&  P(Y_i = y) \, P( \mathcal{L}(i) \mid Y_i = y ) \\
&=& \pi_y \prod_{j \in \mathcal{J}(i)} \alpha_j^{\mathbf{I}[y_i^j = y]} \, (1 - \alpha_j)^{\mathbf{I}[y_i^j \not= y]}. \label{eq:bayesrule}
\end{eqnarray}

Of course, the values of parameters $\pi_y$ and $\alpha_j$ are unknown and must be provided. In Section~\ref{subsec: assessments}, we show how to estimate the $\alpha_j$ from a small set of expertly labeled examples. For $\pi_y$, we simply assume that each class has equal prior probability.
Denoting the estimates as $\hat{\alpha}_j$ and $\hat{\pi}_y$, we can estimate the probability that example $i$ has label $y$ by plugging them into equation~(\ref{eq:bayesrule}).

To predict a hard label for example $i$, we can select $\hat{y}_i = \arg\max_{y} \hat{P}( Y_i = y \mid \mathcal{L}(i) )$ where $\hat{P}(\cdot)$ conveys that we are using the estimated parameters in equation~(\ref{eq:bayesrule}). To predict soft labels, we can use the posterior probabilities from the model, $\hat{P}( Y_i = y \mid \mathcal{L}(i))$. We call the posterior probabilities the confidence scores for the hard, argmax labels.

\subsection{Estimating labeler accuracies}
\label{subsec: assessments}

Estimates of labeler accuracy may already be available from commercial labeling services. If no such data is available, one will need a small collection of known ground truth data denoted $\mathcal{D}_{\text{assess}}$ and have to dedicate a portion of the labeling budget to assessing labelers in order to estimate ${\alpha_j}$. 

In our work, we use the maximum likelihood estimate (MLE) to compute $\hat{\alpha}_j$ for each $j \in \mathcal{J}$ as the proportion of correctly labeled instances by labeler $j$ in the assessment set, separately from the fixed labeling budget:
\begin{equation}
\hat{\alpha}_j = \frac{1}{|\mathcal{D}_{\text{assess}}|} \sum_{i \in \mathcal{D}_{\text{assess}}} \mathbf{I}[Y_i^j = y_i].
\end{equation}

\subsection{Strategy for label collection}
\label{subsec: collecting_labels}
For practical applications under a limited labeling budget, we want to maximize the number of examples that are labeled without compromising quality. To balance this trade-off, we consider two strategies for collecting labels dynamically. In both cases, we assume that we cannot control which labelers label which examples, and that labelers are selected without replacement. 

\paragraph{Strategy 1: Confidence threshold}
\label{subsec:method_confidence_threshold}
We collect labels on a given example until the computed confidence in the aggregate label exceeds a predefined confidence threshold $\tau$, up to a maximum of $\kappa$ labels per example. Practically, $\kappa$ ensures that we do not exhaust the labeling budget on particularly difficult instances. This strategy is a good choice for streaming datasets where not all data points are available at once, or for extremely large datasets where the dataset size is larger than the labeling budget.

\paragraph{Strategy 2: Uncertainty sampling} We use the active learning strategy of uncertainty sampling  \cite{DBLP:journals/corr/LewisG94} to focus labeling effort on the most uncertain or ambiguous examples. Uncertainty $u_i$ is computed based on the confidence in the aggregate label for each example $i$ as follows:
\begin{equation}
    u_i = 1 - \max_y \hat{P}(Y_i = y \mid \mathcal{L}(i)).
\end{equation}
After each label, we update $u_i$ and select the example about which we are most unsure next. We repeat this process until the entire labeling budget is spent. Prior to collecting the first label, each example’s uncertainty is identical and maximum, such that we always collect a single label for all examples first before requesting any additional labels. This strategy is a good choice for traditional datasets that are smaller than the labeling budget. 

We report results from applying each strategy in section \ref{sec: synthetic_experiments}.

\section{Synthetic Data Experiments}
\label{sec: synthetic_experiments}

In order to evaluate the effectiveness of the \ourmethod\ method under controlled conditions, we construct a simulation framework to model possible datasets and labelers. 

\subsection{Label aggregation methods} 
We simulate labeler behavior with the probabilistic model described in section \ref{subsec: imputing_gt} under varying conditions, and compare to three common label aggregation strategies:




\paragraph{Majority voting (MV)} The standard approach where each labeler has an equal vote, and the label with the majority of votes is selected. Confidence in the aggregate label is calculated as the proportion of votes for the selected label over the total number of votes.

\paragraph{Weighted majority voting (WMV)} Each labeler's vote is weighted by their estimated accuracy $\hat{\alpha}_j$. Labelers with higher accuracies have more influence on the final decision. Confidence is calculated as the total weight supporting the selected label divided by the total sum of weights.

\paragraph{Soft voting (SV)} Each labeler's vote is distributed between the two labels based on their estimated accuracy $\hat{\alpha}_j$. Specifically, labeler $j$ contributes a weight of $\hat{\alpha}_j$ to the label they provided and a weight of $1 - \hat{\alpha}_j$ to the other label. The aggregated label is determined by selecting the label with the higher total weight. Confidence is calculated as the total weight supporting the selected label divided by the sum of weights for both labels.

\subsection{Experimental setup} We design experiments to assess the performance of \ourmethod\ versus the three baselines under different labeling budgets and labeler reliability conditions. 

\paragraph{Labeler reliability} We draw an $\alpha_j$ value for each labeler from a uniform distribution over the interval $[a, b]$: $\mathcal{U}(a, b)$. To account for variations in labeler skill and dataset difficulty, we model two different situations. In the first, labelers are more accurate and $\alpha_j \sim \mathcal{U}(0.8, 1.0)$. In the second, labelers are less accurate and $\alpha_j \sim \mathcal{U}(0.6, 0.9)$.

\paragraph{Labeling budget} We report experimental results given the constraint of a fixed labeling budget of $B$ labels collected. In the figures and tables below, sometimes we report results in terms of total labels collected, and elsewhere for clarity sometimes we report results as a function of the average number of labels per example $k=B/N$, where $N$ is the number of labeled examples, which is typically not equal to the size of the dataset. In these cases the budget does not have to be allocated equally to each example. 

\paragraph{Evaluation metrics} We use two metrics to evaluate the performance of the label aggregation methods: 1) the error rate, or proportion of examples where the aggregate label $\hat{y}_i$ does not match the true label $y_i$: $\text{Err} = \frac{1}{N} \sum_{i=1}^N \mathbf{I}(\hat{y}_i \ne y_i)$; and 2) the mean absolute error (MAE), a measure of the average discrepancy between the true label and the method's confidence in that label: $\text{MAE} = \frac{1}{N} \sum_{i=1}^N \left| y_i - \hat{P}(y_i = 1 \mid \mathcal{L}(i)) \right|$. MAE penalizes aggregation methods that are both incorrect and overconfident in their predictions and rewards those providing calibrated confidence values.

\begin{table}[ht]
  \centering
  \begin{subtable}[t]{\linewidth}
    \centering
    \begin{tabular}{lccccc}
      \toprule
      \multicolumn{6}{c}{\textbf{More Accurate Labelers}} \\
      \midrule
      Method & Average $k$ & Best $\tau$ & Labeled Examples (N) & Error Rate & MAE \\ 
      \midrule
      MV   & 5                 & ---  & 3,000 $\pm$ 0.0            & 0.86\% ($\pm$ 0.04\%)          & 10.0\% ($\pm$ 0.17\%) \\
      WMV  & 5                 & ---  & 3,000 $\pm$ 0.0            & \textbf{0.83\% ($\pm$ 0.04\%)} & 9.73\% ($\pm$ 0.17\%) \\
      SV   & 4.78              & 0.99 & 3,140 $\pm$ 20.9           & 0.87\% ($\pm$ 0.04\%)          & 16.8\% ($\pm$ 0.32\%) \\
      GTX  & \textbf{2.75}     & 0.99 & \textbf{4,517 $\pm$ 67.0}  & \textbf{0.83\% ($\pm$ 0.04\%)} & \textbf{1.42\% ($\pm$ 0.06\%)} \\ 
      \bottomrule
    \end{tabular}
    \caption{Comparison for more accurate labelers ($\alpha_j \sim \mathcal{U}(0.8, 1.0)$)}
    \label{tab:ct_experiment_results_a}
  \end{subtable}
  \vspace{2mm}
  
  \begin{subtable}[t]{\linewidth}
    \centering
    \begin{tabular}{lccccc}
      \toprule
      \multicolumn{6}{c}{\textbf{Less Accurate Labelers}} \\
      \midrule
      Method & Average $k$ & Best $\tau$ & Labeled Examples (N) & Error Rate & MAE \\ 
      \midrule
      MV   & 5                 & ---  & 3,000 $\pm$ 0.0             & 10.4\% ($\pm$ 0.30\%)          & 25.1\% ($\pm$ 0.30\%) \\
      WMV  & 5                 & ---  & 3,000 $\pm$ 0.0             & 10.3\% ($\pm$ 0.30\%)          & 24.3\% ($\pm$ 0.30\%) \\
      SV   & 5                 & 0.91 & 3,000 $\pm$ 0.0             & 9.36\% ($\pm$ 0.30\%)          & 36.1\% ($\pm$ 0.30\%) \\
      GTX  & \textbf{4.07}     & 0.96 & \textbf{3,690 $\pm$ 33.0}   & \textbf{9.30\% ($\pm$ 0.30\%)} & \textbf{13.9\% ($\pm$ 0.40\%)} \\ 
      \bottomrule
    \end{tabular}
    \caption{Comparison for less accurate labelers ($\alpha_j \sim \mathcal{U}(0.6, 0.9)$)}
    \label{tab:ct_experiment_results_b}
  \end{subtable}
  \vspace{2mm}
  \caption{\textbf{Comparison of methods using the confidence threshold strategy.} For each method, we report the confidence threshold/fixed number of labels per example that minimizes error over 100 simulations. Results are presented for (a) more accurate labelers and (b) less accurate labelers. See section \ref{subsec:ct_experiment} for details.} 
  \label{tab:ct_experiment_results}
\end{table}

\subsection{\ourmethod\ performance using confidence thresholds}
\label{subsec:ct_experiment}

First, we evaluate our method using the confidence threshold strategy described in section ~\ref{subsec:method_confidence_threshold}. We set a labeling budget of $B = 15000$ labels and initialize ten labelers. We construct a dataset larger than our budget to ensure that we are not data limited. We vary confidence thresholds with a range between 0.85 and 0.99 and a $\kappa=5$ for both \ourmethod\ and SV. For MV and WMV, which give 1.0 confidence after a single label, we collect a constant number of labels for each example, and vary this between 1 and 5. We repeat each experiment 100 times, each time initializing a new dataset and new labelers. For each method, we report results for the confidence threshold that minimizes error (Table \ref{tab:ct_experiment_results}). 

For more accurate labelers (Table \ref{tab:ct_experiment_results_a}), we find that all methods were able to achieve similar error rates. However, our method only required $k = 2.75$ labels per example to achieve this performance, while the next most efficient option, SV, required $k = 4.78$.  This efficiency allows us to label approximately 50\% more data within the same budget. We find that our method is significantly more calibrated as well, with a MAE of $1.42\%$, while the alternatives report MAEs between 10\% and 15\%. 

For less accurate labelers (Table \ref{tab:ct_experiment_results_b}), we again observe that our method achieves a comparable error rate to the alternatives we consider. Our method requires only $k = 4.07$ labels per example compared to $k = 5$ for the other methods. This efficiency allows us to label approximately 23\% more data within the same budget. Moreover, our method demonstrates significantly better calibration, reporting a MAE of 13.9\%, whereas the alternatives have MAEs ranging from 24\% to 36\%. These results suggest that even when labelers are less reliable, our method still enables more data to be labeled within a fixed budget.

We plot our results across all trials for the more accurate labelers in Figure {\ref{fig:c_t_experiment_pareto_line_graph}}. This figure illustrates the relationship between efficiency and error rate, showing that \ourmethod\ requires fewer labels per example to achieve similar error rates compared to the other options.

\begin{table}[ht]
    \begin{tabular}{lcc|cc}
    \toprule
    \multicolumn{1}{c}{} & \multicolumn{2}{c}{\textbf{More Accurate Labelers}} & \multicolumn{2}{c}{\textbf{Less Accurate Labelers}} \\
    \cmidrule(lr){2-3} \cmidrule(lr){4-5}
    \textbf{Method} & Error Rate & MAE & Error Rate & MAE \\ 
    \midrule
    MV   & 2.64\% ($\pm$ 0.43\%)          & 5.22\% ($\pm$ 0.55\%) & 17.9\% ($\pm$ 0.95\%)  & 22.2\% ($\pm$ 0.96\%) \\
    WMV  & 2.70\% ($\pm$ 0.47\%)          & 5.24\% ($\pm$ 0.59\%) & 17.4\% ($\pm$ 0.80\%)  & 22.0\% ($\pm$ 0.84\%) \\
    SV   & 3.85\% ($\pm$ 0.42\%)          & 12.7\% ($\pm$ 1.02\%) & 14.9\% ($\pm$ 0.88\%)  & 32.7\% ($\pm$ 1.02\%) \\
    GTX  & \textbf{0.29\% ($\pm$ 0.07\%)} & \textbf{0.64\% ($\pm$ 0.16\%)} & \textbf{9.46\% ($\pm$ 0.80\%)} & \textbf{16.8\% ($\pm$ 1.25\%)} \\ 
    \bottomrule
  \end{tabular}
  \vspace{2mm}
  \caption{\textbf{Comparison of methods using the uncertainty sampling strategy.} For each method, we report the results over 10 simulations. Results are presented for more accurate labelers ($\alpha_j \sim \mathcal{U}(0.8, 1.0)$) and less accurate labelers ($\alpha_j \sim \mathcal{U}(0.6, 0.9)$). See section \ref{subsec: us_results} for details.}
  \label{tab:us_experiment_results}
\end{table}

\subsection{\ourmethod\ performance using uncertainty sampling}
\label{subsec: us_results}
\begin{figure}[h]
    \centering
    \begin{subfigure}[b]{\linewidth}
        \centering
        \includegraphics[width=\linewidth]{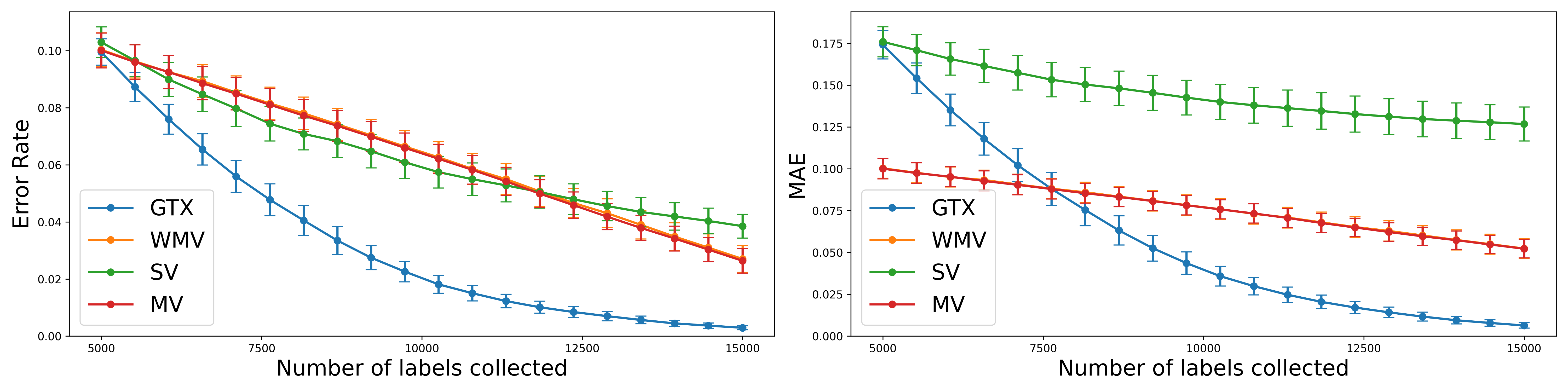}
        \caption{Results for more accurate labelers ($\alpha_j \sim \mathcal{U}(0.8, 1.0)$)}
        \label{fig:u_s_experiment_line_graphs_a}
    \end{subfigure}
    \begin{subfigure}[b]{\linewidth}
        \centering
        \includegraphics[width=\linewidth]{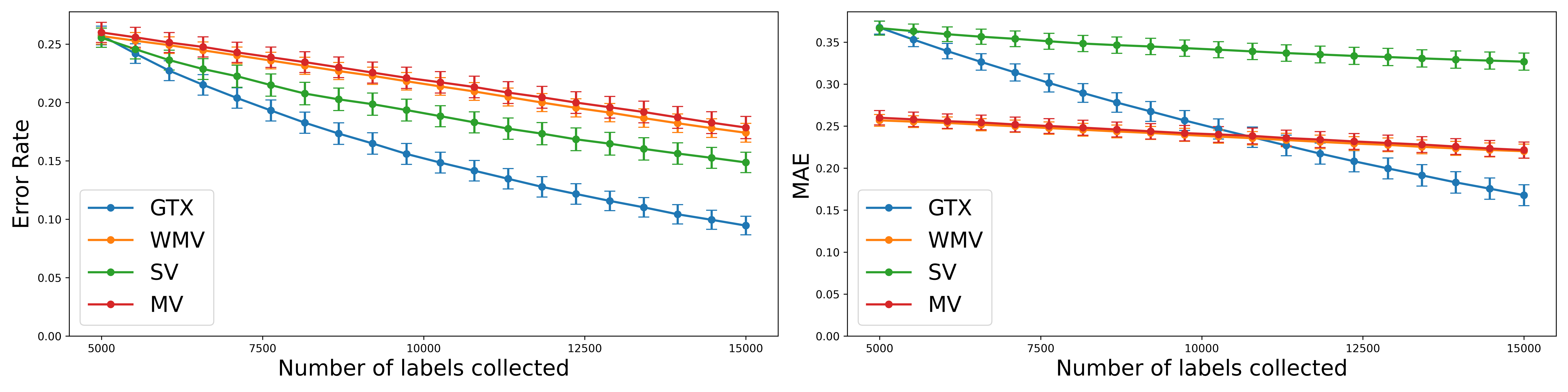}
        \caption{Results for less accurate labelers ($\alpha_j \sim \mathcal{U}(0.6, 0.9)$)}
        \label{fig:u_s_experiment_line_graphs_b}
    \end{subfigure}
    \caption{\textbf{Dynamics of the uncertainty sampling strategy.} For each method, we report how error rate and MAE improve as we collect labels using uncertainty sampling over 10 trials. The dataset has 5000 examples. We plot data starting at 5000 labels collected, after each example is labeled once. The error bars report standard error. Results are presented for (a) more accurate labelers and (b) less accurate labelers. See section \ref{subsec: us_results} for details.}
    \label{fig:u_s_experiment_line_graphs}
\end{figure}

We also evaluate our method using the uncertainty sampling strategy (described in section \ref{subsec:method_confidence_threshold}). We construct a dataset with 5000 examples and initialize 10 labelers. We set a labeling budget of $k = 3$ labels per example, or $B = 15000$ labels in total. We label this dataset with each method using uncertainty sampling, repeating each simulation 10 times. After each label is collected, we record the error rate and MAE across the entire dataset (Figure \ref{fig:u_s_experiment_line_graphs}) in order to visualize the dynamics of each method. We only plot results after each example received an initial label, as the metrics we report are not very meaningful on unlabeled data. For both more accurate and less accurate labelers, we observe that \ourmethod\ has meaningfully lower error rates on average than the most performant alternative throughout the entire iterative process.

With more accurate labelers, we report an average final error rate of $0.29\%$ (Table \ref{tab:us_experiment_results}). For most of the simulation, SV is the next best alternative. However, its performance trails off after 12,500 labels are collected, allowing MV and WMV to outperform it. MV concludes the simulation with the second best performance, an error of $2.64\%$. We notice that the error on this task is correlated with the final MAE. SV performs very poorly on MAE, and we conjecture that this hurts the ability of uncertainty sampling to select informative examples for labeling. With less accurate labelers, we observe that all methods perform worse but \ourmethod\ continues to outperform the baselines. 

\section{Fine-tuning With Imputed Labels}
\label{sec: real_data_experiments}

We demonstrate the performance of \ourmethod\ in generating good imputed labels for fine-tuning through an experiment in image classification.

\begin{figure}
    \centering
    \begin{subfigure}[b]{4.1cm}
        \centering
        \includegraphics[width=4cm]{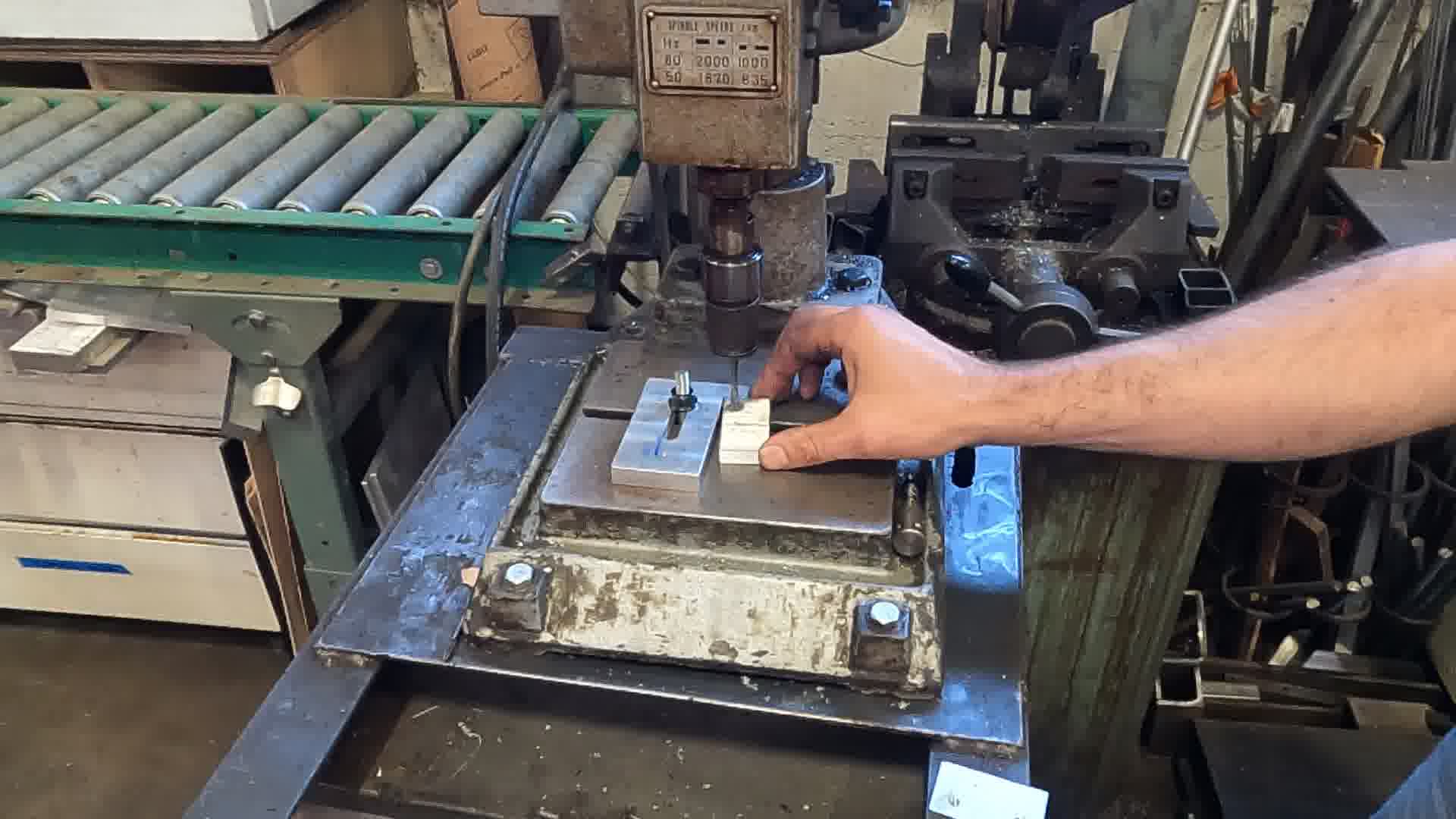}
        \caption{class=YES}
    \end{subfigure}
    \begin{subfigure}[b]{4.1cm}
        \centering
        \includegraphics[width=4cm]{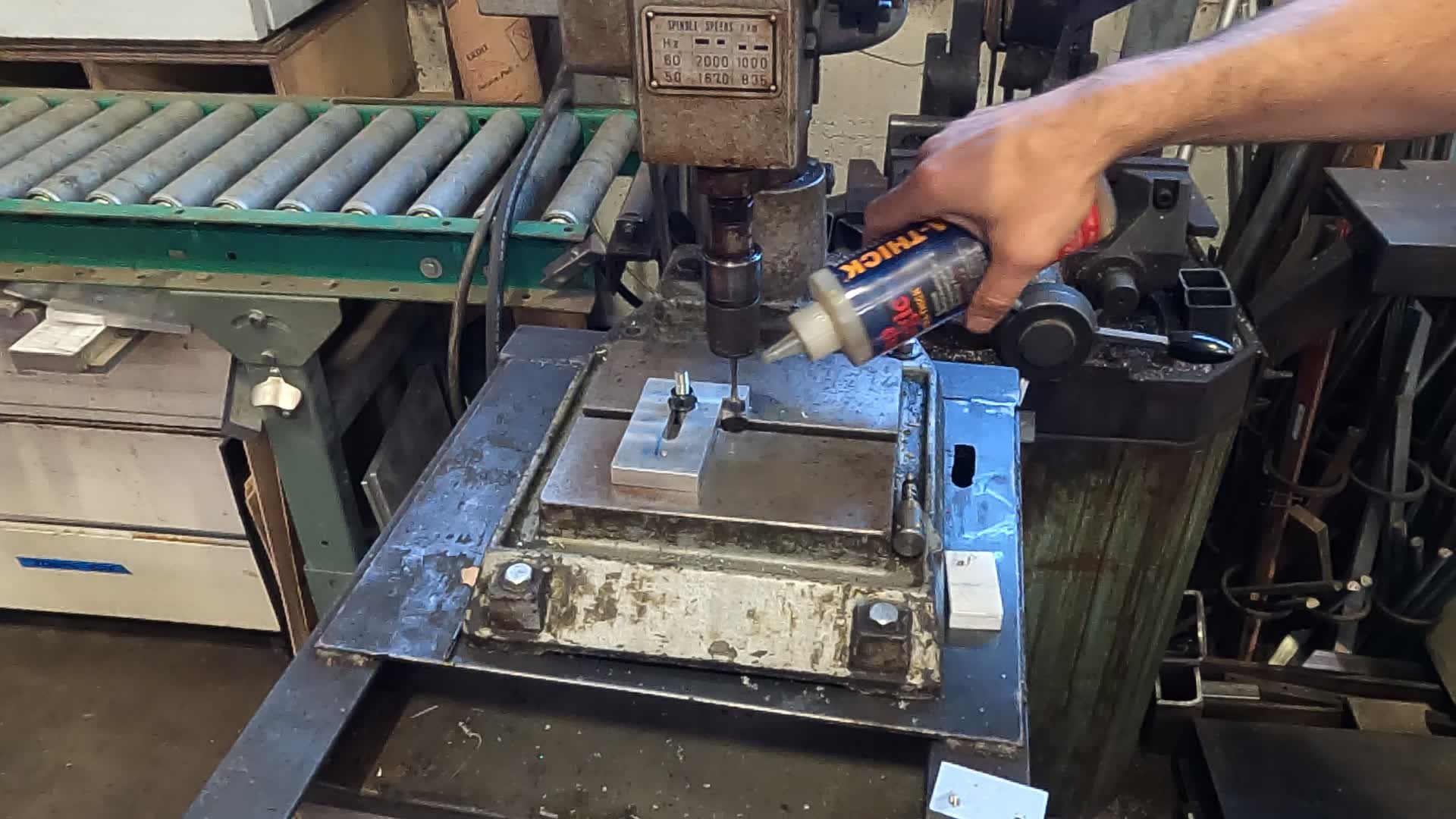}
        \caption{class=NO}
    \end{subfigure}
    \caption{Example YES and NO frames from the singlehole-part-tapping dataset for the question: {\it Is the wood block being drilled correctly? It must be touching the metal guide block on the left and the person must be holding it with his thumb and at least one additional finger.}}
    \label{fig:dataset}
\end{figure}

\paragraph{Dataset} We collect a binary image classification dataset from a video taken in an industrial setting, where an operator is drilling holes in a wooden part. The dataset classifies individual video frames based on whether the operator adheres to best drilling practices as shown in figure \ref{fig:dataset}. There is a total of 1,050 images in this dataset, which we make publicly available at \url{https://huggingface.co/datasets/sunildkumar/singlehole-part-tapping}.

\paragraph{Labeling } We simulate two cohorts of labelers following the same process as in section \ref{sec: synthetic_experiments}: a high-quality cohort ($\alpha_j \sim \mathcal{U}(0.8, 1.0)$) and a lower-quality cohort ($\alpha_j \sim \mathcal{U}(0.6, 0.9)$). We reserve 10\% of the images in the dataset for labeler assessments to estimate individual labeler accuracies, and use those estimates when applying label aggregation methods. We reserve an additional 10\% of the dataset for evaluation. For different fixed labeling budget amounts, so long as there is remaining budget, we collect up to a maximum of $\kappa=5$ labels for each image from these synthetic labelers, and we choose a confidence threshold of $\tau=0.95$ for label aggregation methods requiring a threshold. Once the budget has run out, the set of $N$ labeled images constitutes the training set for our fine-tuning experiments. We repeat the labeling and aggregation for each of the four methods MV, WMV, SV and GTX with 5 different random seeds and labeling budgets ranging from 250 to 4000 labels total.

\paragraph{Training } We fine-tune a pretrained EfficientNet-b0~\cite{DBLP:journals/corr/abs-1905-11946} for 3 epochs on each of the training datasets with imputed labels obtained by each of the four methods, and report its performance on the held out evaluation set. We repeat the training and evaluation 3 times for each of the 5 random seeds. 

\begin{figure}
    \centering
    \begin{subfigure}[b]{\linewidth}
        \centering
        \includegraphics[width=4.6cm]{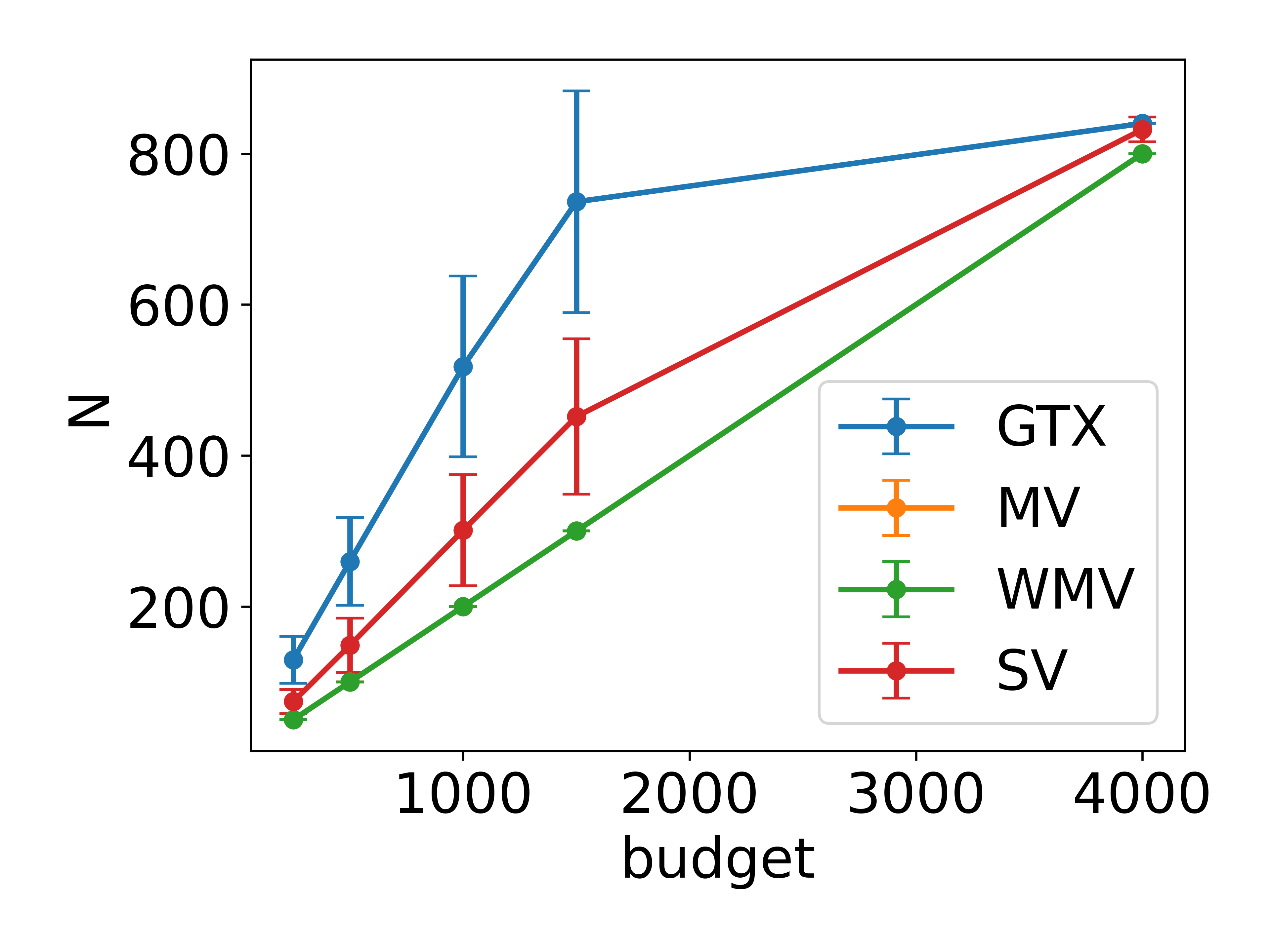}
        \includegraphics[width=4.6cm]{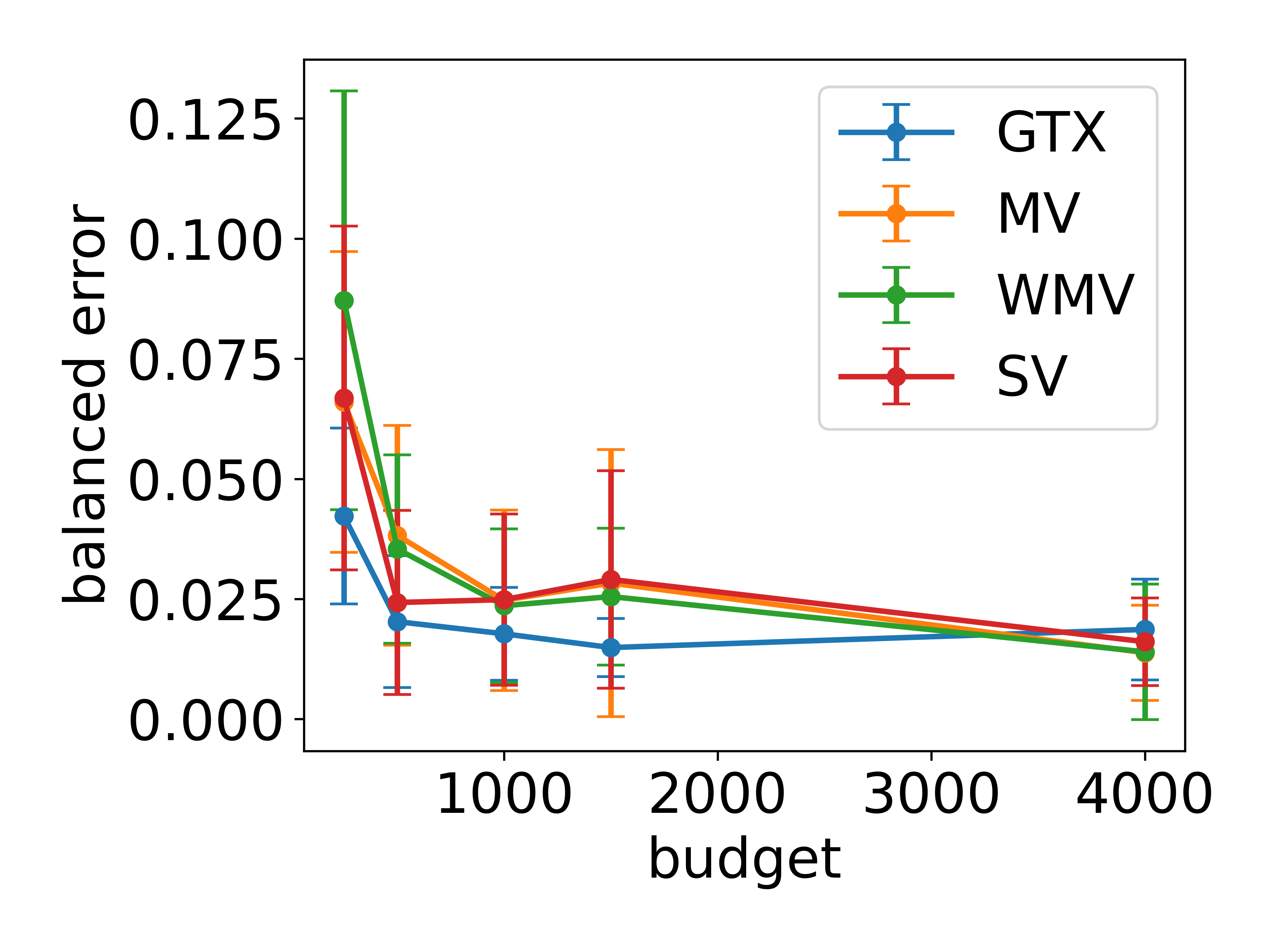}        
        \includegraphics[width=4.6cm]{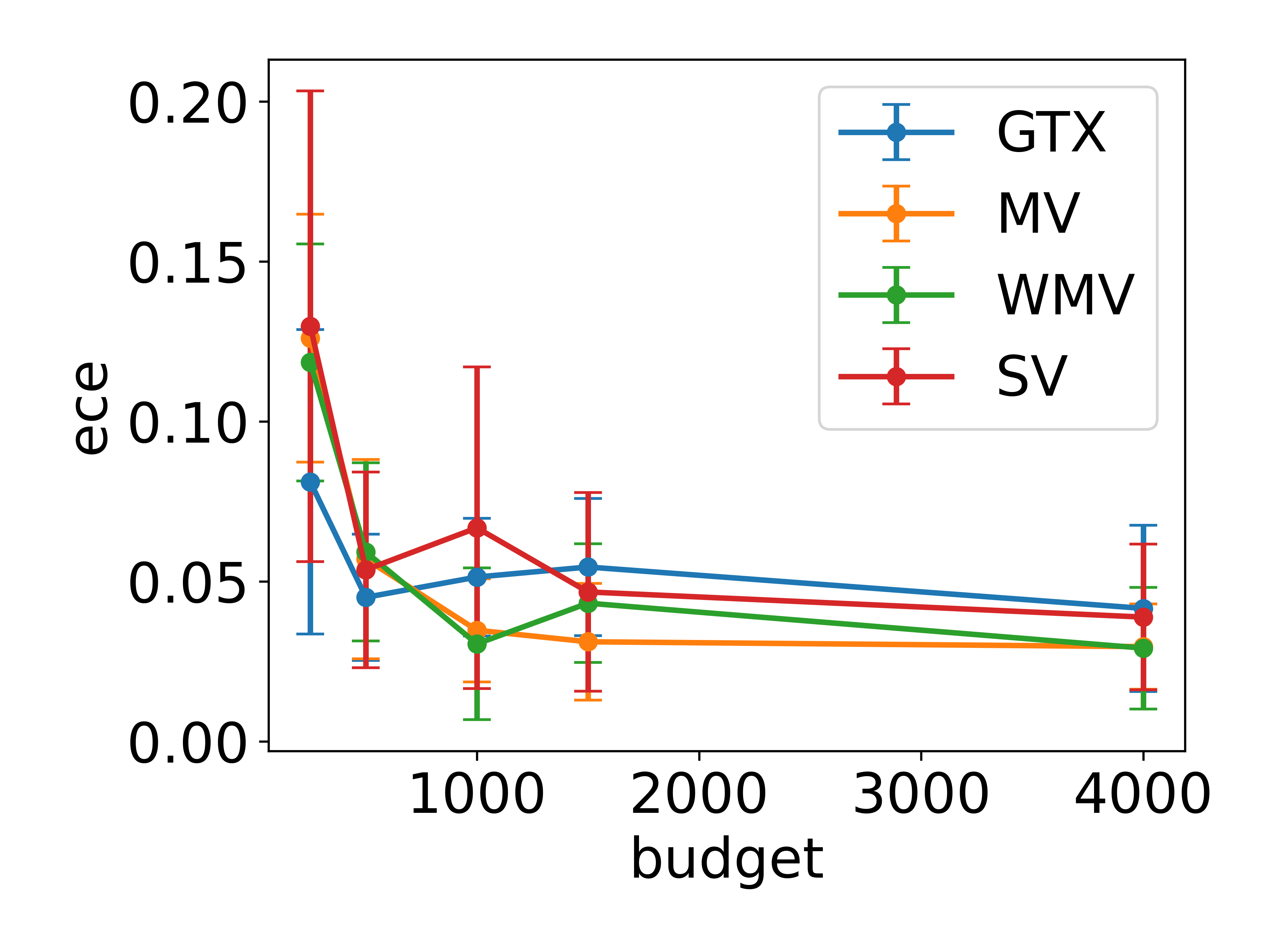}
        \caption{Accurate labelers cohort results ($\alpha_j \sim \mathcal{U}(0.8, 1.0)$)}
    \end{subfigure}
    \begin{subfigure}[b]{\linewidth}
        \centering
        \includegraphics[width=4.6cm]{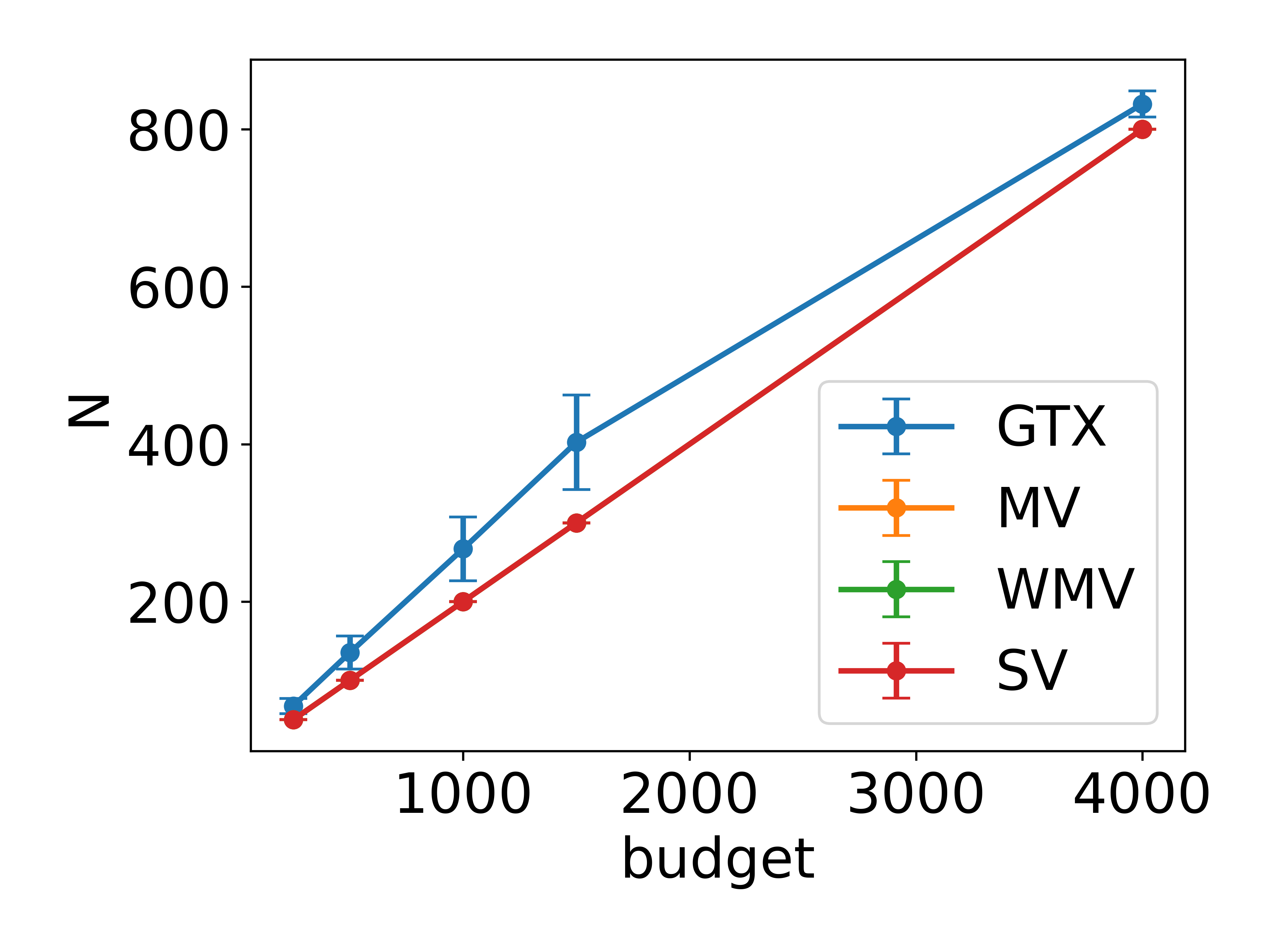}
         \includegraphics[width=4.6cm]{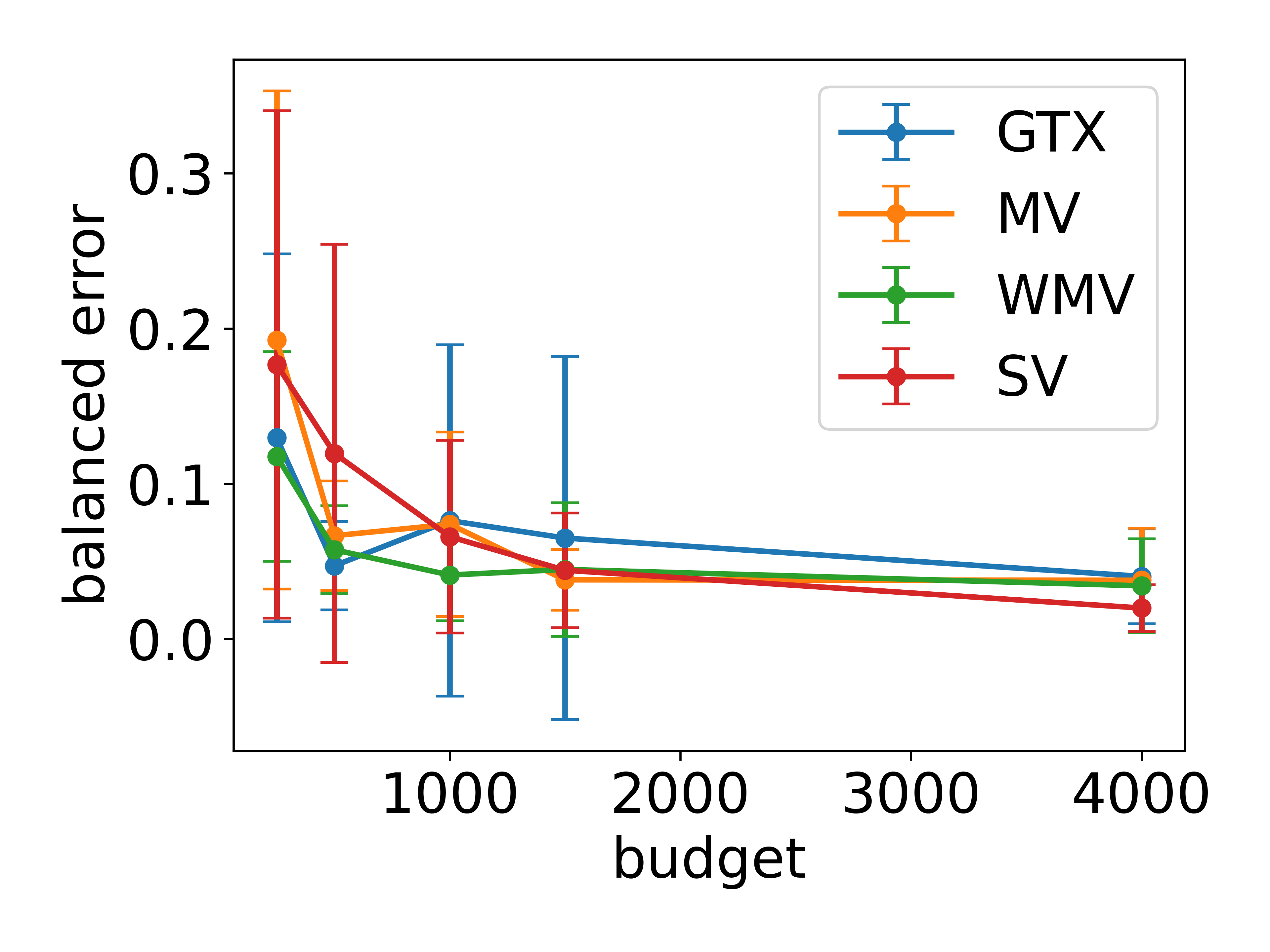}
        \includegraphics[width=4.6cm]{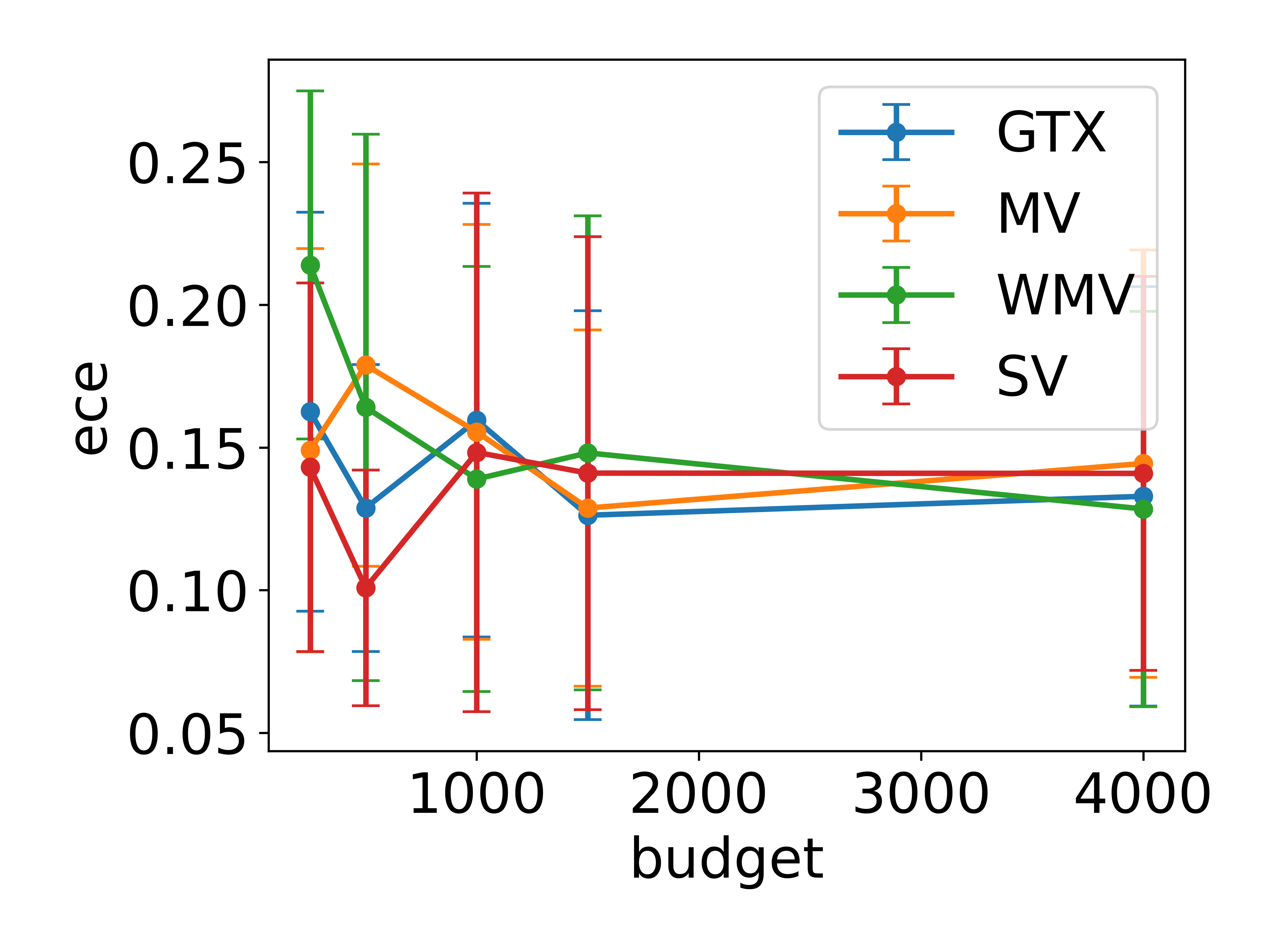}
        \caption{Less accurate labelers cohort results ($\alpha_j \sim \mathcal{U}(0.6, 0.9)$)}
    \end{subfigure}
    \caption{Number of training set examples (leftmost column) and finetuning results as a function of labeling budget for \ourmethod\ compared with the three baselines. We report the balanced error rate (middle column) and expected calibration error (rightmost column) on the withheld evaluation set.}
    \label{fig:finetuning-results}
\end{figure}

Figure \ref{fig:finetuning-results} shows the results. We observe that \ourmethod\ labels significantly more data that the baselines on smaller budgets and with higher-quality labelers. In these situations therefore the resulting models train on a larger dataset and see an increase in performance (lower balanced error) as well as lower expected calibration error. For the accurate labeler cohort, the repeated-measures Friedman's $\chi$-square test on the balanced error reports that labeling methods are different (p-value=2.5e-09), and post-hoc Dunn pairwise analyses shows that \ourmethod's balanced error is significantly lower than MV (p-value=0.043) and WMV (p-value=0.038). In this set of experiments, however, we observe that increasing the labeling budget to the point where most of the available training data points can be labeled by majority vote, or reducing labeler quality to the point where more labels must be collected per example on average by \ourmethod, erases its advantage (see also Appendix \ref{appendix:fine-tuning}). Further analysis and experiments are needed to determine the situations where \ourmethod\ labeling would be more efficient specifically for fine-tuning.

\section{Related Work}

Many authors have studied the problem of training machine learning models on noisy labels~\citep{li2024, baek2023, natarajan2013}, while others have studied the problem of detecting and pruning incorrect labels from a training set~\citep{northcutt2021}. In a related line of work, authors have examined the problem of inferring accurate labels from unreliable, crowd-sourced ones~\citep{dawidskene,raykar2010,muhammadi2013,sheng2019}. In our work, we apply methods from this latter line of research as a means of learning from unreliably labeled data. We also incorporate margin-based sampling from the active learning literature when determining which examples require additional labels and should be prioritized for labeling next~\citep{settles2009}.

\section{Limitations, Extensions, and Conclusion}
\label{sec: discussion}
We have described the \ourmethod~method for collecting and using labels of variable quality in order to obtain a more accurately labeled dataset for fine-tuning ML models on a fixed labeling budget. Here, we discuss the limitations and possible extensions to this approach. 

{\bf Labeler accuracies and assessment strategies} A fundamental assumption of the \ourmethod~method is that we have access to estimates of individual labeler accuracies. Often those can be obtained for returning labelers in the form of benchmark assessments, a version of which we describe in section \ref{subsec: assessments}, or inter-annotator consensus statistics, based on performance on previous datasets and labeling tasks. However, if labeling accuracies are not available, some of the labeling budget will need to be spent on labeler quality assessments before we can benefit from the \ourmethod\ approach. We have shown in section \ref{sec: synthetic_experiments} that \ourmethod\ achieves greater predicted label accuracy than the baselines for a fixed labeling budget, and so we argue that this would be a good investment of a portion of the budget. We leave it to future work to study strategies for selecting the number of assessments needed for a good estimate of labeler accuracy. Here confidence intervals or Bayesian credible intervals could be used.

{\bf Streaming data} Our uncertainty sampling experiments assumed that all unlabeled examples in the dataset are immediately available to use. However, in some settings examples arrive sequentially in a stream (e.g., from a video camera). The \ourmethod\ approach using a confidence threshold strategy is still directly applicable in such a streaming setting, but there would be a smaller pool of examples to solicit labels from for uncertainty sampling (since many examples would have yet to arrive). With a smaller pool of examples to pick from, we should expect the benefits of \ourmethod\ to be less pronounced.

{\bf Modeling labeler error} In this paper, our labeler error model is limited to a single accuracy estimate for each labeler. An obvious extension is to model labeler error with variable error rates based on example class which has been explored previously~\citep{dawidskene}.

{\bf Fine-tuning on estimated labels} Further work is needed to fully determine the conditions under which fine-tuning models specifically benefit from the labeling efficiency of \ourmethod. 

\paragraph{Conclusion} We have proposed a method called Ground Truth Extension (\ourmethod), which outperforms standard aggregation baselines for predicting the true label of an example with several available labels of variable quality. We have shown how to use existing ground truth to estimate individual labeler accuracies, how to decide whether to solicit more labels for a given data point, and how to find the best aggregate label based on a simple probabilistic model of labeler behavior. \ourmethod\ generates more accurate labels than majority-vote baselines. And we show that these more accurate labels produce more accurate fine-tuned models on real image data when labelers have high accuracy. Future work can address method limitations related to assessing and modeling labeler error.



\small

\bibliographystyle{unsrt} 
\bibliography{references} 

\begin{thebibliography}{10}

\bibitem{zhang2021}
Chiyuan Zhang, Samy Bengio, Moritz Hardt, Benjamin Recht, and Oriol Vinyals.
\newblock Understanding deep learning (still) requires rethinking generalization.
\newblock {\em Communications of the ACM}, 64(3):107--115, 2021.

\bibitem{li2024}
Mengting Li and Chuang Zhu.
\newblock Noisy label processing for classification: A survey, 2024.

\bibitem{dawidskene}
Alexander~Philip Dawid and Allan~M Skene.
\newblock Maximum likelihood estimation of observer error-rates using the em algorithm.
\newblock {\em Journal of the Royal Statistical Society: Series C (Applied Statistics)}, 28(1):20--28, 1979.

\bibitem{settles2009}
Burr Settles.
\newblock Active learning literature survey.
\newblock 2009.

\bibitem{ok2016}
Jungseul Ok, Sewoong Oh, Jinwoo Shin, and Yung Yi.
\newblock Optimality of belief propagation for crowdsourced classification.
\newblock In Maria{-}Florina Balcan and Kilian~Q. Weinberger, editors, {\em Proceedings of the 33nd International Conference on Machine Learning, {ICML} 2016, New York City, NY, USA, June 19-24, 2016}, volume~48 of {\em {JMLR} Workshop and Conference Proceedings}, pages 535--544. JMLR.org, 2016.

\bibitem{Khetan2017}
Ashish Khetan, Zachary~C. Lipton, and Animashree Anandkumar.
\newblock Learning from noisy singly-labeled data.
\newblock In {\em 6th International Conference on Learning Representations, {ICLR} 2018, Vancouver, BC, Canada, April 30 - May 3, 2018, Conference Track Proceedings}. OpenReview.net, 2018.

\bibitem{DBLP:journals/corr/LewisG94}
David~D. Lewis and William~A. Gale.
\newblock A sequential algorithm for training text classifiers.
\newblock {\em CoRR}, abs/cmp-lg/9407020, 1994.

\bibitem{DBLP:journals/corr/abs-1905-11946}
Mingxing Tan and Quoc~V. Le.
\newblock Efficientnet: Rethinking model scaling for convolutional neural networks.
\newblock In Kamalika Chaudhuri and Ruslan Salakhutdinov, editors, {\em Proceedings of the 36th International Conference on Machine Learning, {ICML} 2019, 9-15 June 2019, Long Beach, California, {USA}}, volume~97 of {\em Proceedings of Machine Learning Research}, pages 6105--6114. {PMLR}, 2019.

\bibitem{baek2023}
Moseli Mots'oehli and Kyungim Baek.
\newblock Deep active learning in the presence of label noise: A survey, 2023.

\bibitem{natarajan2013}
Nagarajan Natarajan, Inderjit~S. Dhillon, Pradeep Ravikumar, and Ambuj Tewari.
\newblock Learning with noisy labels.
\newblock In Christopher J.~C. Burges, L{\'{e}}on Bottou, Zoubin Ghahramani, and Kilian~Q. Weinberger, editors, {\em Advances in Neural Information Processing Systems 26: 27th Annual Conference on Neural Information Processing Systems 2013. Proceedings of a meeting held December 5-8, 2013, Lake Tahoe, Nevada, United States}, pages 1196--1204, 2013.

\bibitem{northcutt2021}
Curtis Northcutt, Lu~Jiang, and Isaac Chuang.
\newblock Confident learning: Estimating uncertainty in dataset labels.
\newblock {\em Journal of Artificial Intelligence Research}, 70:1373--1411, 2021.

\bibitem{raykar2010}
Vikas~C Raykar, Shipeng Yu, Linda~H Zhao, Gerardo~Hermosillo Valadez, Charles Florin, Luca Bogoni, and Linda Moy.
\newblock Learning from crowds.
\newblock {\em Journal of machine learning research}, 11(4), 2010.

\bibitem{muhammadi2013}
Jafar Muhammadi, Hamid~Reza Rabiee, and Abbas Hosseini.
\newblock Crowd labeling: a survey.
\newblock {\em arXiv preprint arXiv:1301.2774}, 2013.

\bibitem{sheng2019}
Victor~S. Sheng and Jing Zhang.
\newblock Machine learning with crowdsourcing: {A} brief summary of the past research and future directions.
\newblock In {\em The Thirty-Third {AAAI} Conference on Artificial Intelligence, {AAAI} 2019, The Thirty-First Innovative Applications of Artificial Intelligence Conference, {IAAI} 2019, The Ninth {AAAI} Symposium on Educational Advances in Artificial Intelligence, {EAAI} 2019, Honolulu, Hawaii, USA, January 27 - February 1, 2019}, pages 9837--9843. {AAAI} Press, 2019.

\end{thebibliography}


\clearpage
\section*{Supplementary Material}

\begin{appendices}
\section{Fine-tuning on imputed labels}
\label{appendix:fine-tuning}

\begin{figure}[h]
    \centering
    \includegraphics[width=0.4\linewidth]{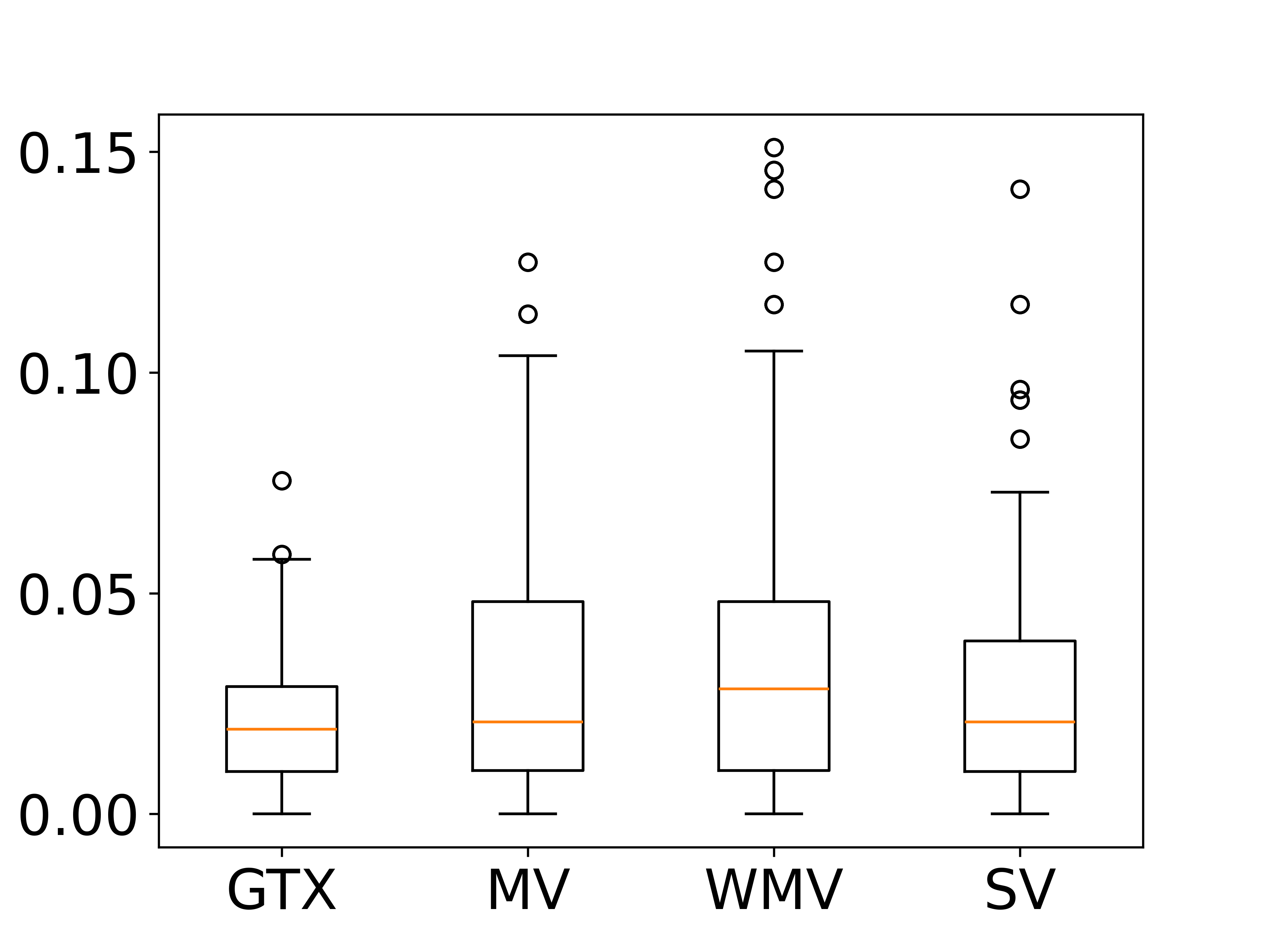}
    \includegraphics[width=0.4\linewidth]{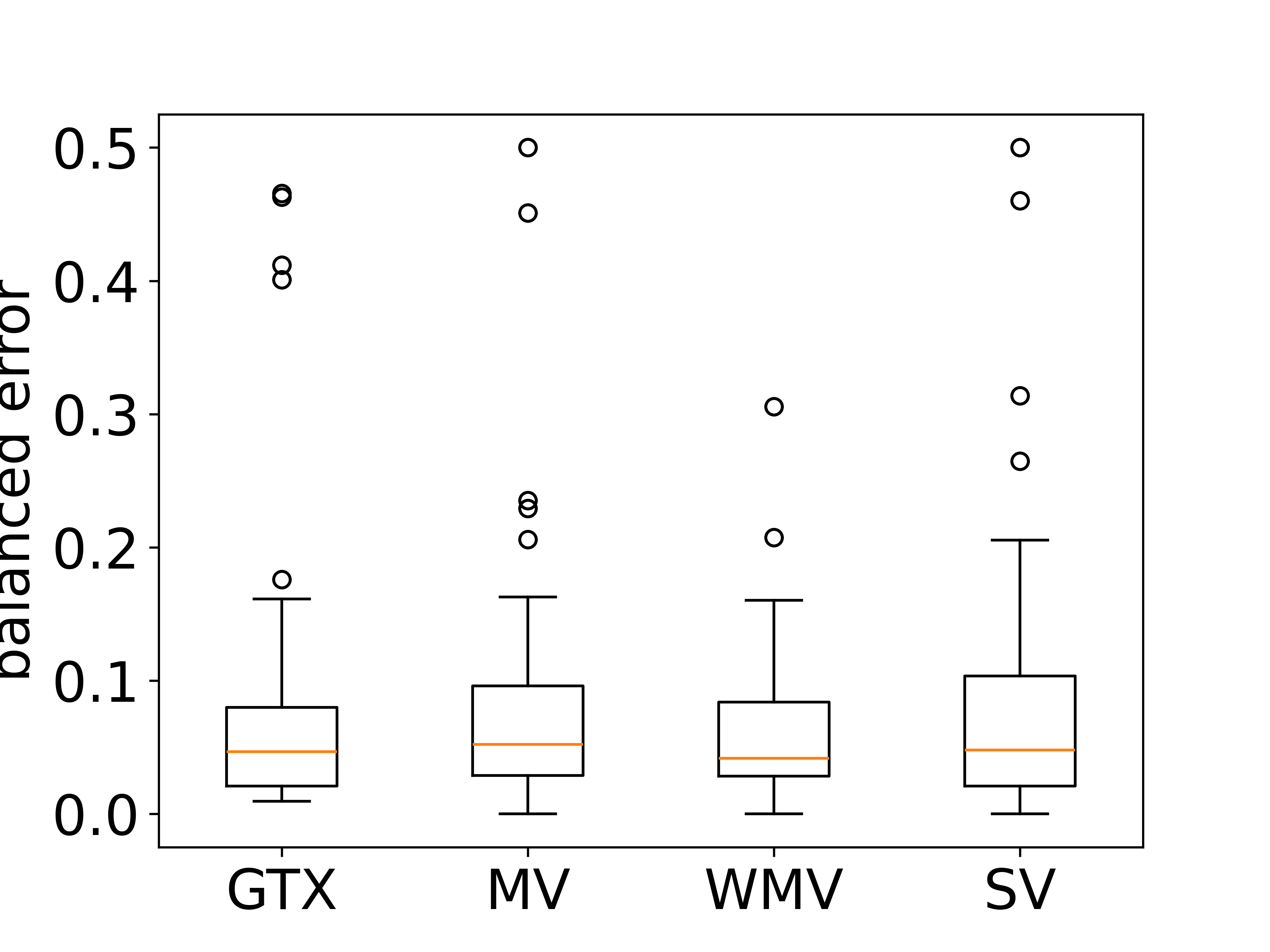}
    \caption{Boxplots of balanced error on the evaluation set after fine-tuning a pretrained EfficientNet-b0 on the singlehole-part-tapping image classification dataset: labels from the accurate labeler cohort (left) vs less accurate labeler cohort (right). Note different $y$ scales. }
    \label{fig:boxplots}
\end{figure}

We find that less accurate labelers erase the fine-tuning advantage of \ourmethod\ over the baselines, as shown in figure \ref{fig:boxplots}. 

\end{appendices}

\end{document}